\documentclass[letterpaper]{article} 
\usepackage[preprint]{aaai24}  
\usepackage{times}  
\usepackage{helvet}  
\usepackage{courier}  
\usepackage[hyphens]{url}  
\usepackage{graphicx} 
\usepackage{natbib}  
\usepackage{caption} 
\usepackage{algorithm}
\usepackage{algorithmic}

\usepackage{newfloat}
\usepackage{listings}
\DeclareCaptionStyle{ruled}{labelfont=normalfont,labelsep=colon,strut=off} 
\floatstyle{ruled}
\newfloat{listing}{tb}{lst}{}
\floatname{listing}{Listing}
\title{Learning a Hierarchical Planner from Humans in Multiple Generations}
\author{
    Leonardo Hernandez Cano\equalcontrib\textsuperscript{\rm 1},
    Yewen Pu\equalcontrib\textsuperscript{\rm 2},
    Robert D. Hawkins\textsuperscript{\rm 3},
    Josh Tenenbaum\textsuperscript{\rm 1},
    Armando Solar-Lezama\textsuperscript{\rm 1}
}
\affiliations{
    \textsuperscript{\rm 1}Massachusetts Institute of Technology,
    \textsuperscript{\rm 2}Autodesk Research,
    \textsuperscript{\rm 3}University of Wisconsin-Madison
}

\usepackage{amsmath}
\usepackage{url}            
\usepackage{booktabs}       
\usepackage{amsfonts}       
\usepackage{nicefrac}       
\usepackage{microtype}      
\usepackage{xcolor}         
\usepackage{graphicx}
\usepackage{caption}
\usepackage{cleveref}

\begin{document}

\maketitle

\newcommand{\Gs}[0]{\mathcal{G}}
\newcommand{\Us}[0]{\mathcal{U}}
\newcommand{\Ss}[0]{\mathcal{S}}
\newcommand{\Ts}[0]{\mathcal{T}}
\newcommand{\M}[0]{\mathfrak{M}}
\newcommand{\A}[0]{\mathbf{A}}
\newcommand{\as}[0]{\mathbf{a}}
\newcommand{\LL}{\mathcal{L}}
\newcommand{\NatP}{\mathcal{P}}
\newcommand{\NP}{\texttt{NP}}
\newcommand{\PLANNING}{\texttt{PLANNING}}
\newcommand{\DS}{\texttt{DS}}
\newcommand{\DP}{\texttt{DP}}
\newcommand{\CL}{\texttt{CraftLite}}

\newcommand{\xxcomment}[4]{\textcolor{#1}{[$^{\textsc{#2}}_{\textsc{#3}}$ #4]}}

\newcommand{\ep}[1]{\xxcomment{red}{E}{P}{#1}}
\newcommand{\lhc}[1]{\xxcomment{teal}{L}{H\llap{C}}{#1}}
\newcommand{\rh}[1]{\xxcomment{orange}{R}{H}{#1}}
\newcommand{\asolar}[1]{\xxcomment{blue}{A}{S}{#1}}
\newcommand{\todo}[1]{\xxcomment{orange}{TO}{DO}{#1}}

\makeatletter
\newcommand{\REPEATN}[1]{\ALC@it\algorithmicrepeat%
\ #1 \textbf{times}\begin{ALC@rpt}}%
\newcommand{\ENDREPEAT}{\end{ALC@rpt}\ALC@it\algorithmicend}%
\makeatother

\makeatletter
\newcommand{\CASE}[2]{\ALC@it\textbf{case}%
\ #1 \textbf{is} #2\begin{ALC@rpt}}%
\newcommand{\ENDCASE}{\end{ALC@rpt}\ALC@it\algorithmicend}%
\makeatother

\begin{abstract}
A typical way in which a machine acquires knowledge from humans is by programming.
Compared to learning from demonstrations or experiences, programmatic learning allows the machine to acquire a novel skill as soon as the program is written, and, by building a library of programs, a machine can quickly learn how to perform complex tasks. 
However, as programs often take their execution contexts for granted, they are brittle when the contexts change, making it difficult to adapt complex programs to new contexts.
We present natural programming, a library learning system that combines programmatic learning with a hierarchical planner.
Natural programming maintains a library of decompositions, consisting of a goal, a linguistic description of how this goal decompose into sub-goals, and a concrete instance of its decomposition into sub-goals.
A user teaches the system via curriculum building, by identifying a challenging yet not impossible goal along with linguistic hints on how this goal may be decomposed into sub-goals.
The system solves for the goal via hierarchical planning, using the linguistic hints to guide its probability distribution in proposing the right plans.
The system learns from this interaction by adding newly found decompositions in the successful search into its library.
Simulated studies and a human experiment (n=360) on a controlled environment demonstrate that natural programming can robustly compose programs learned from different users and contexts, adapting faster and solving more complex tasks when compared to programmatic baselines.

\end{abstract}

\section{Introduction}


A hallmark of a good learner is the ability to accumulate skills learned from different teachers.
Crucial to this process is the ability to apply a skill learned in one context to a different context.
For instance, consider a cooking robot that learns to cook from different people. 
Since not all kitchens have the same set of ingredients, the robot must adapt its recipes by substituting the ingredients when needed.
Specifically, we are interested in building a learning agent that interacts with a sequence of users $u_1 \dots u_n$ on a sequence of related (yet distinct) Markov Decision Processes (MDPs) $M_1 \dots M_n$. 
A capable agent should be able to acquire skills from a particular user $u_i$ in a particular context $M_i$, and adapt these skills to different MDPs. 
By doing so, the agent is able to robustly compose skills learned from different teachers despite context changes, allowing it to solve longer, more complex tasks the more users it interacts with.

How do we build such an agent? 
Central to the ability to adapt is the \textit{representation} of the learned skills. 
Deep Reinforcement Learning \cite{guez2019investigation,mnih2013playing} stores skills implicitly as weights of a Neural Network, and requires large amounts of data to learn a new skill, making it infeasible for real-time learning from people.
On the other hand, one might represent skills as programs with crisp and modular semantics, allowing the agent to learn quickly from users \cite{bunel2018leveraging}.
As programs are inherently compositional, a user can quickly build complex skills into the agent by combining smaller programs into bigger ones.
Yet, the brittleness of programs when contexts change is a known problem \cite{sumers2022talk, bonawitz2011double}, making it difficult to apply to our setting, where the dynamics of the MDPs change.
Hierarchical Planning approaches \cite{kaelbling_hierarchical_2011, konidaris2018skills,silver_learning_2021,devin_learning_2017}, on the other hand, find ways to achieve a given goal at run-time, and therefore adapt to changing contexts by-design. 
In this work, we ask: how might we build a system where a group of naive users can teach a hierarchical planner of novel goals and decompositions?

\begin{figure}
\includegraphics[width=\linewidth]{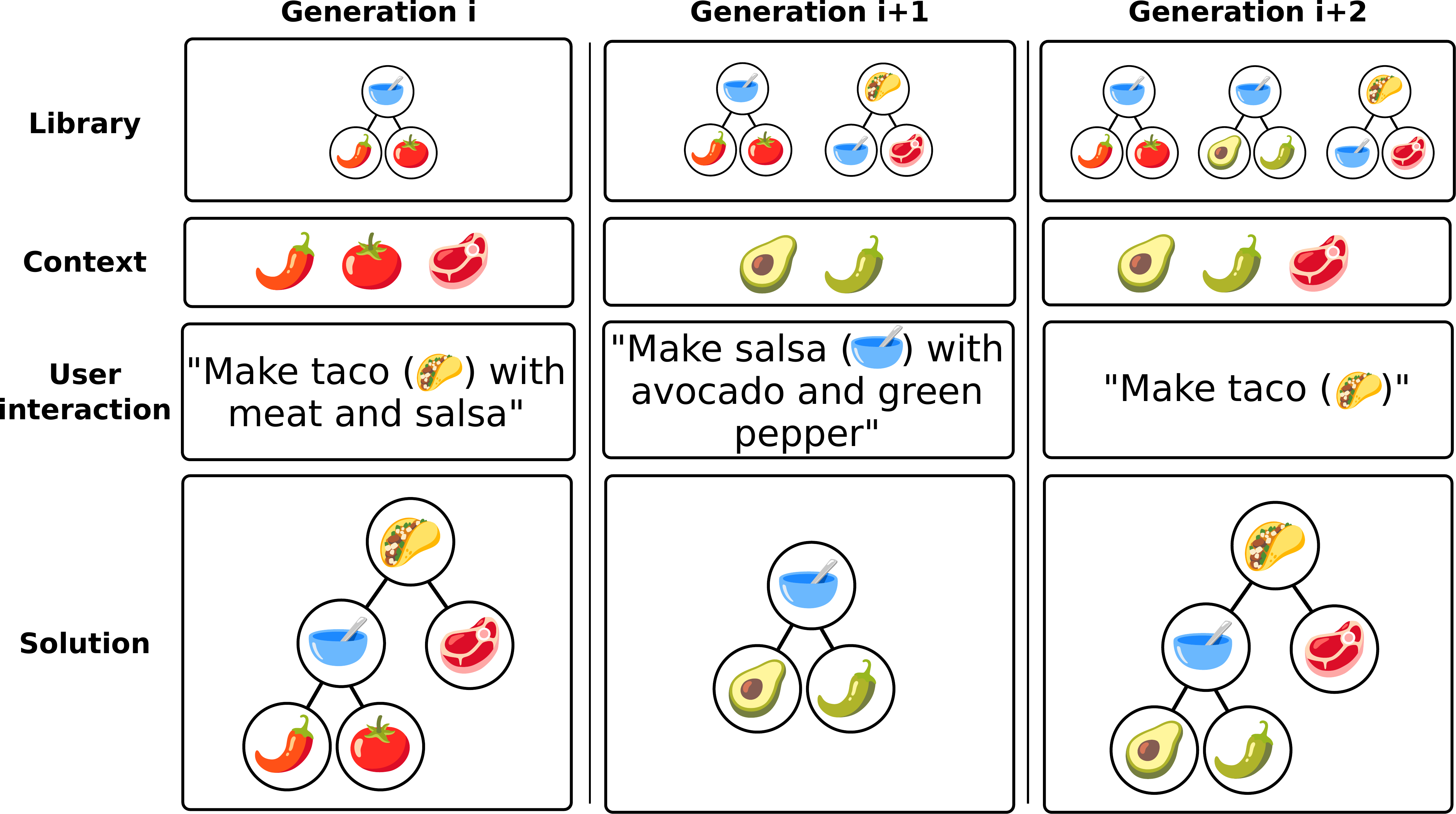}
\caption{Overview of a Natural Programming system. \NP\, grows a library (top) of decompositions, consisting of how goals, such as a bowl of salsa, decompose into sub-goals, such as a red pepper and a tomato. Users have different contexts (mid), shown as different sets of starting ingredients in different kitchens. Users interact with \NP\ by providing goals and linguistic hints on how to decompose them into sub-goals.
\NP\ solves the user-given goals guided by the linguistic hints in the form of full plans (bot), and adds new decompositions into its library to be used (and adapted) in the future.}
\label{fig:tacos}
\end{figure}
To that end, we propose \textit{Natural Programming} (\NP), a hierarchical planner that learns from human interactions. 
Internally, \NP\, maintains a library of decompositions we termed \emph{natural programs}, consisting of a goal, a linguistic hint on how it may decompose into sub-goals, and a concrete instantiation of this decomposition into actual known sub-goals.
Users inform \NP\, of \emph{new goals}, along with linguistic hints on how these new goals may be decomposed into sub-goals the system might already know.
\NP\, then solves for this goal, using the linguistic hints as guide to propose likely decompositions.
When a goal is satisfied by a plan, \NP\, learns the newly discovered decompositions by adding a natural program into its library. 
Similar to a programming system, a user can quickly teach \NP\, a new skill and have it be used immediately.
Unlike a programming system, the naive user does not need to explicitly specify the decomposition of a goal into sub-goals, but rather only linguistic hints on how a goal \emph{might} decompose into sub-goals -- absolving the responsibility of having to provide a correct decomposition.
Finally, since \NP\, is fundamentally a hierarchical planner, it adapts across different contexts by-design.
competitive Thus, similar to the program synthesis paradigm, \NP\, can take in a search problem and find a novel solution for the current context. However, unlike traditional library-building program synthesis approaches, where search is performed on top of a library of deterministic programs, \NP\, searches for a solution recursively over the library of hierarchical search problems, in the style of hierarchical planning \cite{kaelbling_hierarchical_2011, konidaris2018skills,silver_learning_2021,devin_learning_2017}. This allows the system to reuse known solutions to a given problem, but also to adapt known solutions to the current context at run-time. 
Evaluation of \NP\, against competitive baselines on both simulation and a large-scale user study ($n=360$, $\text{type} = \text{crowd-workers}$, $\text{total time}=\text{90 hours}$) demonstrates that \NP\ learns best from different users and contexts, compared with programming baselines~\footnote{We will also open source the \CL{} environment, the NP system and all the collected user-interaction dataset.}. 

\paragraph{Contributions}
This work makes 3 contributions: \textbf{Formalism} --defining the problem of Generational Learning under changing Contexts (GLC). \textbf{System} --Developing \NP, the first system (to our knowledge) that combines a hierarchical planner and generational human learning, allowing naive users to teach new goals and decompositions to the planner. \textbf{Evaluation} -- Developing \CL{}, a simple yet fully fledged GLC problem, where we validate that \NP\ performs well compared to programmatic baselines.

\section{Motivating Example: Making Tacos}


Imagine being tasked with building a cooking robot. The robot will be deployed to 3 different kitchens, learning to cook while interacting with a human. Suppose the robot knows from previous interactions how to cook salsa with tomato and red pepper. \textbf{Kitchen1:} In the first user's kitchen there was a red pepper, a tomato, and meat. The first user asked the robot to cook a taco. The robot did not know how to do this. So, the first user explained to the robot that tacos are cooked by combining salsa and meat. The robot knows how to cook salsa, so it proceeds to cook a taco by first cooking salsa and then combining the salsa with the meat. After this interaction, the robot knows that tacos are cooked by combining salsa and meat. \textbf{Kitchen2:} The robot then goes to another kitchen, where the second user asks it to cook salsa. Instead of a red pepper and a tomato, there is avocado and green pepper. So the second user guides it to prepare the new salsa. After interacting with the second user, the robot knows 2 ways of making salsa. \textbf{Kitchen3:} The robot then goes with the third user, who asks it to cook a taco. The robot adapts its knowledge of how to cook tacos --learned from the first user--, and how to cook salsa with avocado and green pepper --learned from the second user--, to successfully make a taco in the third kitchen. See Figure~\ref{fig:tacos}.

\paragraph{Properties of GLC}
Problems like this one can be formalized as sequential decision-making problems with (1) a hierarchical structure (opportunities to decompose and re-use acquired knowledge), (2) a need for rapid adaptation (to enable real-time user interactions and to solve the issue that past skills may be invalid in the current context), and (3) linguistic guidance (allowing naive users to provide information useful to derive new solutions). In this paper, we study how to design a learning agent that can efficiently tackle these problems by exploiting these properties.
\section{Problem Statement}
\label{sec:gpc}






We now define the Generational Learning under Changing Contexts (GLC) problem. We define this to study agents that learn from users in a generational setting. Specifically, we are interested in agents that learn from different teachers consecutively, each teacher within a context consisting of its own goals and dynamics.

Let $S$ and $A$ be the set of states and the set of actions, respectively. An MDP is a tuple $(S, A, s, t, R)$, where $s \in S$ is the starting state, $t: S \times A \to S$ is a transition function and $R: S \times A \to \mathbb{R}$ is the reward function. A goal $g$ is a predicate function over trajectories $g([s_1, \ldots, s_n]) \in \{0,1\}$, where $s_i \in S$. This work considers the setting where the set of goals, $G$, is finite, and dynamics are deterministic.

\paragraph{CMDP}
A \textbf{context} is a tuple $(g,s,t)$, consisting of a goal $g \in G$, a starting state $s \in S$, and a transition function $t: S \times A \to S$. A Contextual MDP (CMDP) ~\cite{hallak2015contextual} maps a context $c$ into a specific MDP $\M(c) = (S,A,s,t,R_g)$, where $s, t , R_g$ are the starting state, transition function, and goal-conditioned reward function, respectively.

\paragraph{GLC}
A GLC problem is a tuple $(\Us, \Gs, \Ss, \Ts, \M, F)$, consisting of sequences (representing different generations) of \textbf{users} $\Us = (u_1 \dots u_n)$, \textbf{goals} $\Gs = (g_1 \dots g_n)$,  \textbf{starting states} $\Ss = (s_1 \dots s_n)$, \textbf{dynamics} $\Ts = (t_1 \dots t_n)$, a \textbf{CMDP} $\M$, and set of \textbf{final states} $F \subseteq S$. The CMDP $\M$ assigns a generation-specific starting context $c_i = (g_i, s_i, t_i)$ to the MDP $(S,A,s_i,t_i,R_{g_i})$. In this work, $S$ and $A$ are fixed across generations, while $s_i, t_i, R_{g_i}$ change across generations. We model users as mapping the context of a current state and goal and transition function into a query that contains a natural language hint, $u : (g,s,t) \rightarrow (g, \mathit{hint})$.

Agents act over the generations of a GLC problem consecutively. During generation $i$, the agent acts in $\M(c_i)$ until reaching a final state, optionally interacting with user $u_i$ in each step. Once a final state is reached, the agent proceeds to the next generation. The agent is allowed to leverage all the experiences (the specifics will differ from agent to agent) from previous generations to act in the current generation. Agents are equipped with a model of the environment which they can query \emph{as a black box}\footnote{i.e. the dynamics of $\M(c)$ are hidden from the agent, but it can take actions and observe outcomes ``in simulation''.} to plan their actions. Acting on each generation results in a generation-specific cumulative reward $r_i$. The objective of the GLC problem is to maximize the global cumulative reward $\sum_i r_i$.

\section{Background}

We now describe few standard approaches to build an agents that solves GLC problems. These approaches serve as background to our proposed approach, Natural Programming.

\subsection{Planning}

In planning systems, the behavior of an agent is specified by providing a goal $g \in G$. The planner then leverages its model of the environment to search for a sequence of actions such that the given goal is met. Once such a sequence is found, each action in the sequence is executed on the environment, which updates the current context with the resulting state. Planning allows users to specify only desired outcomes, which reduces the burden on users and places it on the planning system. This also allows agents to adapt to new contexts automatically by generating new plans. However, the space of plans and the way in which goals should decompose into suitable sub-goals are typically given as part of the planning problem~\cite{kaelbling_hierarchical_2011, arumugam2017accurately, arumugam2019grounding} instead of being user-given.


\subsection{Direct Programming (\DP)}
In \DP, the behavior of an agent is specified through programs. Programs are sequences of actions which are to be taken on the environment.
For this work, programs do not have variables and control flows\footnote{Think of a system of macros.}.
A community of users teach \DP\ new skills by adding programs into its library.
Users interact with \DP\ by reasoning over their current context and library, and writes an appropriate program for \DP\ to run.

\paragraph{Program Definition}
Programs are defined by providing a mapping $\mathit{name}_{\textit{new}} \rightarrow [\mathit{name}_1 \dots \mathit{name}_k]$. This mapping is then added to the library.

\paragraph{Execution}
Once a program has been defined, it can be executed under the current context $c$ by providing its $\mathit{name}$. To do this, \DP\, performs a lookup for $L[\mathit{name}]$ under the current library $L$, resulting in a sequence of primitive actions.
The state of the current context is updated after executing actions on the environment.


\paragraph{Generational Library Learning}
\DP\, starts a GLC problem with the library of primitive actions $L_1 = L_A = \{\operatorname{name}(a_i) \to [a_i]\}_{\forall a_i \in A \}}$, where all the actions in $A$ are referred to by their names.
\DP\, grows the library as users add new functions to it. More precisely, the library at the start of the $i+1$ generation is $L_{i+1} = L_{i} \cup L_{i}^{\textit{new}}$, where $L_{i}^{\textit{new}}$ are the mappings created by $u_i$.

\paragraph{Brittleness of Functions}
As the starting states and transition functions in a GLC changes over generations, a function written under one context often breaks in a new context. This brittleness prevents an user from re-using and building upon functions written by other users.
\subsection{Direct Synthesis (\DS)}
\DS\, augments \DP\, with language-guided program synthesis, where the system searches the library for a sequence of programs that solve a given goal.
Thus, \DS\, maintains a library of functions just like \DP, and, in addition, \DS\, is made aware of the set of goals $G$ and can query the environment $\M(c)$ as a black box.

\paragraph{Programming with Search Problems}
When interacting with an agent to specify its behavior in \DS, users provide a \textbf{search problem} $(g, \mathit{hint})$, which contains a goal $g \in G$ and a natural language string $\mathit{hint}$ on how to achieve it. Here, instead of browsing a library like in \DP, users provide search problems solely by reasoning over their current context.

\paragraph{Execution via Search}
\DS\, searches for a program such that the induced sequence of actions satisfy the given search problem when carried out in $\M(c)$.
In a typical program synthesis fashion \cite{devin_learning_2017,ellis2020dreamcoder}, \DS\, performs rejection sampling\footnote{Most program synthesis approaches adopt the generate and check framework.}. Like most program synthesis with library learning frameworks, \DS\, searches through compositions of programs from the library (see Algorithm~\ref{alg:DS}). This allows \DS\, to find increasingly complex behaviors within a fixed search budget as the library grows. Note that programs in \DS\, are deterministic.

\begin{algorithm}[tb]
\caption{DS Execution}
\label{alg:DS}
\textbf{Input}: $(g, \mathit{hint})$, $\M(c)$, $L$\\
\textbf{Output}: Program $f$
\begin{algorithmic}[1] 
\FOR{$L^i \in \{L^1, L^2, \ldots, L^{\textit{max\_len}}\}$}
    \REPEATN{$\mathit{n\_iter}$}
        \STATE $(f_1, \ldots, f_i) \sim \operatorname{propose}( - | (g, \mathit{hint}), L^i)$
        \STATE $\mathbf{a} = \operatorname{map}(\operatorname{flatten}, (f_1, \ldots, f_i))$
        \IF{$g(\M(c)(\mathbf{a}))$}
            \STATE $f = \mathit{hint} \to [f_1, \ldots, f_i]$
            \RETURN $f$
        \ENDIF
    \ENDREPEAT
\ENDFOR
\RETURN fail
\end{algorithmic}
\end{algorithm}

\paragraph{Propose} The search is guided by a $\operatorname{propose}$ function in the manner of language-guided synthesis~\cite{wong2021leveraging, li2022competition}, where, depending on the linguistic hint, certain compositions of programs become more likely to be sampled. This is crucial, as the search space of $L^i$ grows exponentially as $i$ increases. The difficulty of having a good propose function is \emph{a lack of data}. Prior approaches, such as~\cite{wong2021leveraging, li2022competition, suhr2019executing,wang2015building}, have required significant labeling efforts of paired instances of natural-language to programs (in our case, program decompositions). Instead, we turn to pretrained LLMs as a source of prior knowledge. Specifically, we consider two ``backends'' of our $\operatorname{propose}$ function:

\begin{small}
\begin{align*}
            &  \operatorname{propose}_{\textit{sim}}(f_1 \dots f_i | (g, \mathit{hint}), L^i) \\
    \propto & \|\phi_{\textit{LLM}}(f_1.\textit{name} + \dots + f_i.\textit{name}) - \phi_{\textit{LLM}}(\mathit{hint})\| \\
            & \operatorname{propose}_{\textit{prompt}}(f_1 \dots f_i | (g, \mathit{hint}), L^i) \\
    =       & \texttt{LLM}(\texttt{prompt}[\mathit{hint},L^i]),
\end{align*}
\end{small}

where $\phi$ returns the embedding of a string into a vector space with an LLM, and $\texttt{LLM}$ samples problem decompositions from an LLM. In summary,
$\operatorname{propose}_{\textit{sim}}$ ranks the sequence $(f_1 \dots f_i)$ based on how semantically similar the $\mathit{hint}$ and the concatenation of the program names are; and $\operatorname{propose}_{\textit{prompt}}$ samples sequences of program names from an instruction-tuned LLM, constructing a prompt with the hint and a relevant subset of the library's content.
Details can be found in the Appendix.
While the optimal form of $\operatorname{propose}$ is ultimately domain-specific, we highlight that $\operatorname{propose}$ is compatible by-design with LLM-guided search techniques, which have shown promise in a wide range of domains, including robotics~\cite{vemprala2023chatgpt}.

\paragraph{Generational Library Learning}
\DS\, starts with a library of primitives $L_1 = L_A$.
Given a search problem $(g, \mathit{hint})$, DS produces a program $f$ when $g(\M(c)(f))$, which is then added to the library.
Doing so achieves \textit{compression} \cite{ellis2020dreamcoder} --assigning to a single name a complex composition of programs that is difficult to find--.

\paragraph{Brittleness of Programs}
Note that because dynamics change across contexts, a program $f$ synthesized under a particular context $c_1$ which satisfies a goal $g$ is unlikely to satisfy the same goal under a different context $c_2$, i.e. $g(\M(c_1)(f)) \neq g(\M(c_2)(f))$. As a result, when contexts change, many of the programs in the library $L$ no longer satisfy the goals they are written for.
\section{Natural Programming (\NP)}
\label{sec:np}

Note that \DP\, and \DS\, are able to accumulate knowledge by building a library of programs which can be composed with each other, but are very brittle. On the other hand, note that planning systems are robust because they can regenerate different plans for different contexts. \NP\ seeks to combine the complementary strengths of planning and programming, by maintaining a library of hierarchical search problems learned from naive user interactions. 

In Natural Programming (\NP), ``natural programs'' in the library $\LL$ are mappings from search problems to problem decompositions --sequences of primitives or other search problems--. Crucially, we allow mappings to be non-deterministic. Thus, at execution time and like a planning system, a given search problem can be mapped non-deterministically to a decomposition that is appropriate for the current context, and this happens recursively for each sub-problem in the decomposition. As the library grows, execution becomes faster by leveraging the decompositions in the library.

The crucial difference between \NP\, and \DS\, is that run-time semantics of a program in \DS\, are deterministic. Thus, even if \DS\, can find novel compositions of programs, \DS\, is incapable of recursively leveraging ``alternative implementations'' like \NP\, does.

\paragraph{Programming with Search Problems}
In \NP, like in \DS, users program an agent by providing a \textbf{search problem} $(g, \mathit{hint})$, where $g$ and $\mathit{hint}$ are a goal and natural language hint, respectively.

\paragraph{Execution via Recursive Decomposition}
Given a search problem $(g, \mathit{hint})$, \NP\, searches for a satisfying plan in the manner of hierarchical planning, similar to the \texttt{MAXQ} algorithm~\cite{dietterich2000hierarchical}. We use $\M(c)(\mathbf{a})$ to denote taking an action sequence $\mathbf{a}$ over the contextual MDP defined by context $c$. In summary, to execute an \NP\, is to non-deterministically choose, through search, a decomposition of the given search problem into a sequence of steps, each step being either another search problem or an action primitive. Each step is then recursively executed until a complete search tree which grounds to primitive actions is built (see Algorithm~\ref{alg:NP}). The search is incremental, where \NP\, attempts to first find a single-line decomposition that satisfies the goal $g$, and when that fails, expands the search to $\LL^2$, and so on, where $\LL^i$ stands for the cross-product of $\LL$ with itself $i$ times, including primitives (i.e., sequences of length $i$ of natural programs in $\LL$ or primitives). See Algorithm~\ref{alg:NP}.

\begin{algorithm}[tb]
\caption{NP Execution}
\label{alg:NP}
\textbf{Input}: $(g, \mathit{hint})$, $\M(c)$, $\LL$\\
\textbf{Output}: Search tree $n$
\begin{algorithmic}[1] 

\FOR{$L^i \in \{\LL^1, \LL^2, \ldots, \LL^{\textit{max\_len}}\}$}
    \REPEATN{$\mathit{n\_iter}$}
        \STATE $(n_1, \ldots, n_i) \sim \operatorname{propose}( - | (g, \mathit{hint}), \LL^i)$
        
        \STATE $\mathit{steps} = \operatorname{list}()$
        \STATE $(\_, s', \_) = c$
        \FOR{$n\in \{n_1, \ldots, n_i\}$}
            \STATE $c' = (g, s', t)$
            \STATE $n' =  \operatorname{NP}(n, \M(c'), \LL)$
            \STATE $\mathbf{a'} = \operatorname{flatten}(n')$
            \STATE $s' = \M(c')(\mathbf{a'})$
            \STATE $\mathit{steps}.\operatorname{append}(n')$
        \ENDFOR
    
        \STATE $n = (g, \mathit{hint}) \to \mathit{steps}$
        \STATE $\mathbf{a} = \operatorname{flatten}(n)$
        \IF{$g(\M(c)(\mathbf{a}))$}
            \RETURN $n$
        \ENDIF
    \ENDREPEAT
\ENDFOR
\RETURN fail

\end{algorithmic}
\end{algorithm}

The most salient aspect of the \NP\, execution is the recursive call to \NP\, itself.
This allows \NP\, to recursively try different decompositions until a satisfying sequence is found. 
In practice, instead of using rejection sampling, we implement Algorithm \ref{alg:NP} with a priority queue and caching to avoid redundant executions (see Appendix).

\paragraph{Propose} Both \DS\, and \NP\, share the same $\operatorname{propose}$ implementation, but in the case of \NP, it is sequences of search problems (tuples with a goal and a natural language hint) that are sampled, instead of sequences of program names.

\paragraph{Generational Library Learning}
When Algorithm \ref{alg:NP} succeeds, \NP\, produces a complete \emph{search tree} consisting of the correctly chosen decompositions. Each decomposition maps a search problem into a sequence of either sub-problems or primitive actions.
Each of these mappings are then added to $\LL$, making each of them available for future non-deterministic executions. More precisely, entire search trees are never stored in the library, rather, the library only contains the ``subtrees of height 2'' which describe the decompositions used in all previous successful search trees.


\section{\CL}
To study the effects of systems in the context of GLC, we introduce an environment, \CL, with the following desiderata: the domain should (1) be able to be setup as, and feature the main challenges of the GLC problem, (2) allow crowd workers to consistently learn how to ``program'' agents in under 5 minutes --so a large-scale user study would not be prohibitively expensive--, and (3) have a real-time ($\sim$2s) responsiveness to accommodate for end-user interactions.

\subsection{\CL{} as a Conditional MDP}

\paragraph{State} A state consists of an inventory of multiple items, two crafting input slots, and a single crafting output. There are 29 total possible items in \CL{}.

\paragraph{Action} There are only two kinds of \textbf{actions}, \texttt{input\_x} where x is an item name, and \texttt{craft}. \texttt{input\_x} moves an existing item in the inventory to the input slots, and \texttt{craft} moves the transformed item in the output slot back to the inventory, consuming the inputs. There are 30 actions.

\paragraph{Dynamics} Dynamics are dictated by a \texttt{recipe} book, which specifies which pair of input items can be successfully transformed into an output item. Out of the 29 items, 4 are ``raw materials'' which cannot be crafted, and 25 are ``craftable items'' each with two possible crafting rules. A recipe book is \emph{randomly generated} by choosing one version of the crafting rules for each item--there are $2^{25}$ possible recipe books--. Depending on the particular recipe book, the most complex item can take up to 87 action steps. Thus, the raw command sequence complexity of \CL{} is $30^{87}$.

\paragraph{Goal} Goals consist of a list of \texttt{goal items}. Crafting a new copy of an item in the goal generates a reward of +1.

\subsection{The GLC Problem using \CL{}}
\begin{figure}
\includegraphics[width=\linewidth]{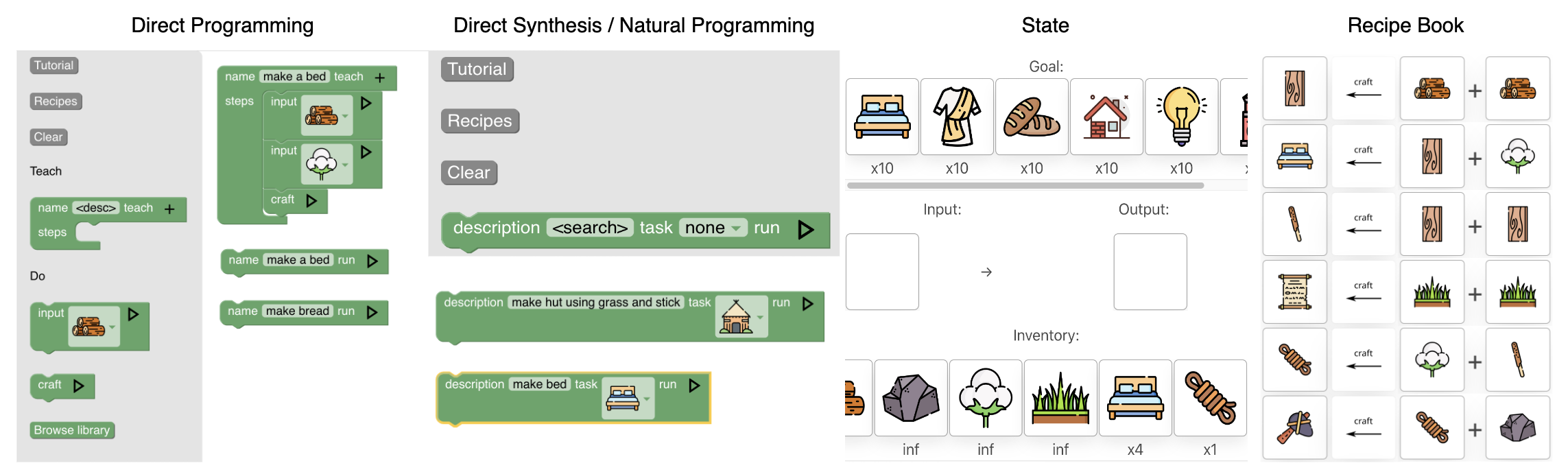}
\caption{The \CL{} UI. The \DP\, programming system allows the user to manually define new functions, and browse a library of existing functions. \DS\, and \NP, systems only allow users to provide search problems. The state shows an inventory of current items and a list of goal items to be completed. Recipe book encodes the generation-specific dynamics, which the user can browse.}
\label{fig:ui}
\end{figure}

At generation $i$, the \textbf{context} $c_i = (g_i, s_0, t_i)$ consists of a list of goal items $g_i = (g_i^1 \dots g_i^r)$, the starting state $s_i$, and the dynamics $t_i$ induced by a random recipe book. In 10 minutes, the user $u_i$ interacts with the learning agent to craft as many goal items as possible under the generation-specific MDP $\M(c_i)$.
The user interface consists of a left programmatic panel based on Blockly \cite{8120404}, and a right panel showing the current state (Figure \ref{fig:ui}). 
\section{Experiments}
We seek to answer the following research questions. 
\textbf{RQ1}: Is \NP\, effective at solving the GPC problems in \CL{}? \textbf{RQ2}: Does the capability of \NP\, improve at a faster rate across generations? \textbf{RQ3}: Does \NP\, allows the user to achieve more tasks with fewer efforts?
In each experiment, users are organized into ``cultural chains''\cite{tamariz2016cultural,tessler2021growing}. Each ``chain'' is a GLC problem as described above, where a sequence of users collaborates with the same learning agent. Learning agents start each chain with library $L_1$. The library is updated after every interaction with a user, and persists to the next generation. 
Chains are carefully matched to use the same random seed at each generation across the chains, thus each generation shares goals and books across different programming systems.

\subsection{Simulated Study}

\begin{figure*}[t]
\begin{center}
\centerline{\includegraphics[width=0.93\linewidth]{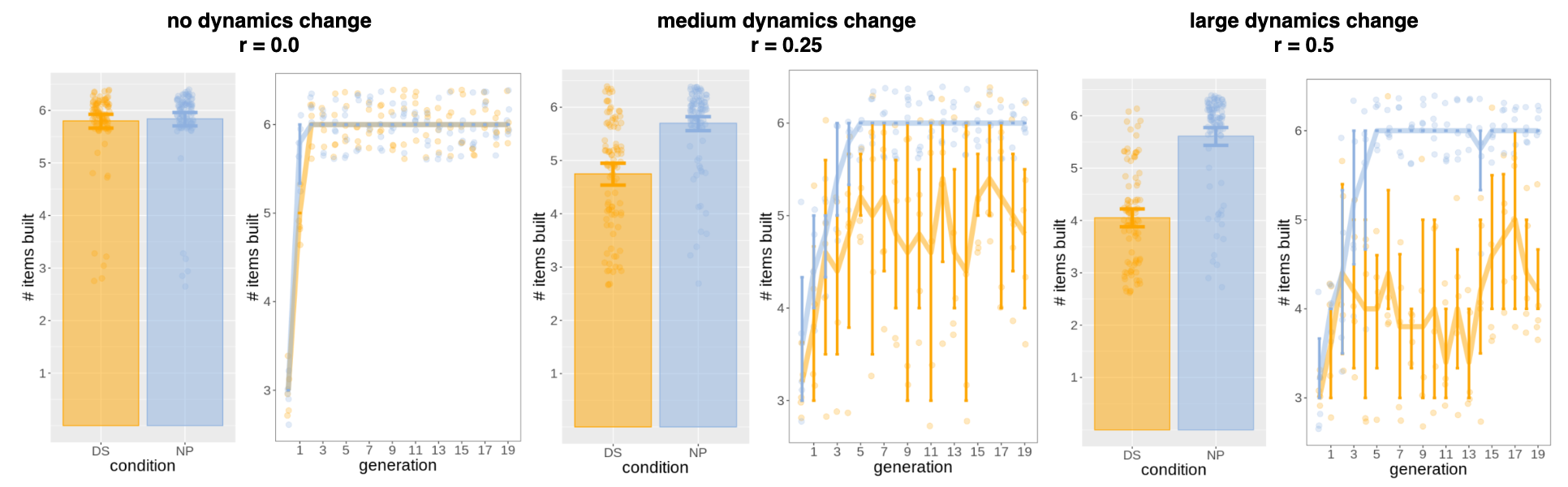}}
\caption{Simulation on how different programming systems perform under different amount of dynamic changes across generations. Total of 600 idealized simulated users. $r = 0.0$ no dynamic change and $r=0.5$ most dynamic change. Error bars are 95\% CIs (nboot=1000), dots represent individual sessions outcomes.
}
\label{fig:simulation}
\end{center}
\end{figure*}


We conduct a simulation study with a large number of simulated users to study the relative effectiveness of \NP\, vs. \DS\, as a function of changing dynamics. 

\paragraph{Varying Dynamics} In \CL{}, every craftable item has 2 possible rules. We can control how much dynamics can vary from one generation to next by adjusting the probability $r$, of how likely the second rule is chosen. A $r=0$ would cause the generated recipe to only contain the first rule for every item, i.e. \emph{no} dynamic changes across generations, while a $r=0.5$ will uniformly sample one of the $2^{25}$ possible recipe books --the most difficult setup possible, with \emph{the most change} in dynamics across generations--.

\paragraph{Batches} At every generation, the context contains (1) a set of 6 randomly chosen ``leaf items'' --craftable items that are not required to craft other items-- were used as goals, (2) the starting state consisting of the 4 raw materials, and (3) a randomly generated recipe book. 
A batch consists of 20 generations of 2 conditions, \DS\, vs. \NP, where the same generation shares the context as described above.
For each value of $r \in [0, 0.25, 0.5]$, we generate 5 batches.

\paragraph{Simulated Users} We simulate users interacting with \DS\, and \NP\,. The simulated users will attempt first to craft a goal item, and wait some amount of time for the solver to succeed. If the solver succeeds withing the allotted time, the user moves onto the next item. Otherwise, the user attempts to recursively craft a prerequisite item (i.e., an item that is used in the recipe of the attempted item). We do not simulate a user for \DP, instead defer evaluation of \DP\, to a real human experiment. Each simulated session has a 2-minute timeout limit, and we set the solver timeout to 10 seconds.

\paragraph{Results} 
\textbf{RQ1}: When dynamics are kept constant ($r=0.0$), \DS\, and \NP\, performs identically on \CL{}. However, as dynamics vary more across generations ($r=0.25$, $r=0.50$), the performance of \NP\, is superior to that of \DS\, (Figure \ref{fig:simulation}). \textbf{RQ2}: We find that \NP\, improves at a faster rate across generations compared to \DS\, under more dynamics variations. \textbf{RQ3}: Because these are simulated users, we defer the measurement of user efforts to a human experiment.

\paragraph{Propose Back-end} We found that for \CL{}, even if we set a lenient solver timeout of 10 seconds (enough for the LLM prompt back-end to provide a response), the similarity-based proposer is generally faster in finding solutions. A detailed comparison between $\operatorname{propose}_\textit{sim}$ and $\operatorname{propose}_\textit{prompt}$ implementations can be found in the Appendix. 
\subsection{Human Experiment}

\begin{figure*}[t]
\begin{center}
\centerline{\includegraphics[width=0.9\linewidth]{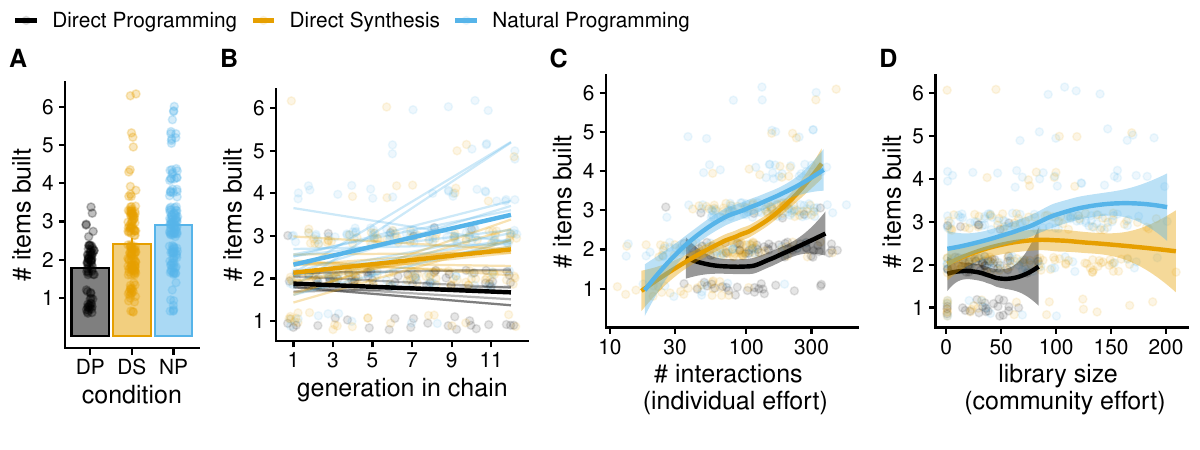}}
\caption{Natural programming (\NP) enables (A) more items to be built overall, (B) improves significantly as successive generations interact with the system, and (C-D) reduces the effort required to reach the same performance. 
Error bars are 95\% CIs; low-transparency dots represent individual sessions; low-transparency lines represent regression fits for individual chains.
}
\label{fig:results}
\end{center}
\end{figure*}

\paragraph{Procedure}  We recruited 360 participants from the Prolific crowdsourcing platform. Participants were recruited from the U.K. and the U.S., excluding people without English proficiency. We paid an average of \$12.06 USD per hour, including bonuses for \$0.4 USD per ``goal''. 
In total, we ran 360 sessions across 12 batches of paired contexts, and of a length of 12 generations. 6 batches containing all three conditions (\DP, \DS, \NP) and 6 only containing the two of greatest interest (\DS, \NP).
Each session is 10 minutes. 
The \DS\, and \NP\, solver is set to have a maximum timeout of 30 seconds, but the user can cancel the solver at any point.

\subsubsection{Results}
\paragraph{RQ1: More items are built overall with \NP}

Across all batches and generations, we find that NP allows people to build the most goal items overall (Figure \ref{fig:results}A; \DP\, mean $= 1.7$ items, $95\% CI = [1.6, 1.9]$, \DS\, mean $= 2.4$ items, $95\% CI = [2.3, 2.6]$, \NP, mean $= 2.9$ items, $95\% CI=[2.7, 3.1]$). 
Both synthesis-based systems perform significantly better overall than \DP\, ($t(71)=5.6, p <0.001$ for \DP\, vs. \DS; $t(71) = 7.9, p<0.001$ for \DP\, vs. \NP, paired).
Compared to \DS, \NP\, produced significantly more items ($t(143) = 4.7, p<0.001$).

\paragraph{RQ2: \NP\, chains improve more rapidly}
Next, we consider how \NP\, improves as additional users in a chain interact with the system (Figure \ref{fig:results}B).
We fit a Bayesian mixed-effects linear regression model predicting the number of items built as a function of the generation in the chain (integer from 1 to 12), the programming system (categorical, \NP\, vs. \DS\, vs. \DP\,), and corresponding interaction terms. 
First, examining the \NP\, condition alone, we found that performance improved significantly across generations, ($b = 6.24, 95\%$ credible interval $= [3.3, 9.3]$). 
We found that this slope was meaningfully larger than the \DS\, condition (diff = $+3.4$, $95\%$ credible interval $= [-0.5, 7.1])$ and the \DP\, condition (diff $= 8.6$, $95\%$ credible interval $= [3.5, 13.5]$).

\paragraph{RQ3: \NP\, requires less effort}

What properties of \NP\, enable these performance benefits? We know it is not due to the difference in user interface, as \DS\, and \NP\, use the same interface. 
We argue instead that \NP\, reduces the amount of effort required to obtain the same results.
We tested this effect by running Bayesian mixed-effects regressions predicting the number of items built as a function of effort and the programming system being used.
We consider two different metrics of effort.
First, we examine \emph{individual effort}, the (log) number of ``submissions'' --registered every time a user attempts to execute a program-- made by a user to the programming system (Figure \ref{fig:results}C).
At a given level of effort, we found that participants in the \NP\, condition were able to craft significantly more items than those in \DP\, condition (diff $=1.25$ items, $95\% CI=[0.8, 1.7]$) or the \DS\, condition (diff $=0.42$ items, $95\% CI=[0.2, 0.6]$). 
Similar results were found for \textit{collective effort} --the total library size (i.e., unique and successful previous interactions) that has accumulated at the given point in the chain (Figure \ref{fig:results}D)--.
\section{Related Works}


\paragraph{Programs as Policies and Planners}
Programs are a staple representation for robot policies. Compared to end-to-end policies, programmatic policies  \cite{andreas2017modular, yang2021program} are more generalizable \cite{inala2020synthesizing, trivedi2021learning}, interpretable \cite{zhan2020learning,bastani2018verifiable}, and easier for humans to communicate \cite{bunel2018leveraging}.
Likewise, it is customary for planners \cite{pmlr-v80-shiarlis18a,konidaris2018skills,silver2021learning} to adopt a programmatic representation. With the advancement of LLMs, recent works have increasingly embraced a programmatic representation in policy \cite{liang2022code, volum2022craft} and
planning \cite{huang2022language, huang2023grounded, silver2022pddl}, taking advantage of LLM's ability to translate between natural languages and programs.

\paragraph{RL and Planning with Changing Dynamics}
Robust operation of agents under changing dynamics \cite{hallak2015contextual, 10.1145/3459991, choi2000reinforcement, xie2022robust} is of crucial importance.
Methods such as \cite{banerjee2017quickest, mesesan2019dynamic,lin2019adaptive} calibrate a learned policy to adapt to novel environments, while \cite{ kaelbling1993learning, squire2015grounding, bodik_programming_2010} maintains a hierarchy of goal representations, allowing the agent to re-plan as environments change. Our work is most similar to \cite{fu2019language,arumugam2017accurately, arumugam2019grounding}, where the human instructions are grounded in \emph{goals} rather than \emph{policies}. Task structure has been leveraged in learning-enabled (neural) planning approaches~\cite{pmlr-v80-shiarlis18a}.

\paragraph{Library Learning}
In library learning, a system continually grows a library of programs, becoming more competent in solving complex tasks over time.
The library learning can be self-driven \cite{ellis2020dreamcoder}, guided by language \cite{wong2021leveraging}, or in direct interactions with a community of users \cite{wang2017naturalizing, wang2016learning,itsblockslol,karamcheti2020learning}. 



\paragraph{Overall} Compared to works in robust policy and planning, we focus on \emph{interactive learning from users} as a mean of acquiring new abstractions and search strategies. Compared to works in library learning, we focus on \emph{goal driven task decompositions} to account for \emph{changing contexts}. Task-oriented dialogue systems \cite{suhr2019executing, fast2018iris, wang2015building} do not perform library learning \footnote{i.e. these systems cannot acquire higher level programs over time}.

\section{Conclusion and Discussion}
We define a new class of problems, GLC, to study how an agent can learn from different teachers and contexts in generations. We developed \NP\, a hierarchical planner that learns new abstractions and decomposition rules from users. We demonstrate, through a large user study ($n=360$), that naive crowd workers can effectively guide \NP\ to solve planning tasks in a space of $30^{87}$ configurations within 12 generations of interactions.

The most direct future work will be scaling to a richer domain consisting of a rich goal space not limited to the fixed set of goal items in \CL{}.
Developing a system that can align user-given goals with how a system interprets these goals \cite{gabriel2020artificial} over a complex goal space is an open, yet exciting field of research.
To scale \NP\ to this richer domain, the planning language must be richer, with variables and control-flow. Studying how users interact with this richer language is also an exciting future direction.

\bibliography{legacy}

\newpage
\appendix

\section{The \textit{propose} function}
\label{sec:propose}

Here we explain in more detail the \textit{propose} function. In Algorithm~\ref{alg:NP}, there is a single propose function being used throughout the execution of an NP. However, we found that the execution of NP has fundamentally two stages. On the one hand, the natural language hint given by the user is very informative for a top-level decomposition of intent. On the other hand, given a good top-level decomposition candidate, it is necessary to recursively leverage previous successful searches in the library. These two stages thus pose fundamentally different challenges: the top-level decomposition benefits from understanding the intent and the context in which the NP was called, whereas the recursive call requires data structures to search the library very efficiently.

Thus, our implementation of \textit{propose} is split into two functions, each tackling one of the two challenges we described above. The \textbf{outer propose} leverages the user utterance, which is stating facts about how the task should be performed (e.g., how the recipe has to be crafted), and thus leveraging language is useful. We implement this as described in the main text. However, language provides no benefits for the \textbf{inner propose} because as context changes, the natural language hints become less relevant, and therefore the inner propose only needs to consider goals. To avoid the performance penalty of querying the LLM, We implement the inner propose as a simpler special case of the outer propose.

\subsection{Inner propose}
\label{sec:inner_propose}
To implement the inner propose, we maintain a data structure that allows fast lookups of all the decompositions in the library associated with any given goal. This focuses the search on the known solutions for a particular goal. Note that each solution is recursively a sequence of Natural Programs which have to be solved. The data structure also allows the proposer to efficiently assign to each decomposition a frequency score and a recency score. These scores are weighted (0.2 and 0.8 respectively) to assign a final score. The weights of these two scores were determined with a small grid search and manual tuning. As a final consideration, the proposer takes in the context (effectively a call-stack) in which it is called, which is used to avoid the possibility of infinite recursion. See Listing~\ref{lst:inner_propose} for the pseudocode of the inner propose.

\paragraph{Frequency score}{This is defined as the number of times the decomposition appears in the library, normalized to 0 and 1 for a given set of candidate decompositions. Frequency helps the system identify which decompositions are more reusable. To see why this is useful throughout the GPC problem, consider that the planner can find implementations that attain a particular goal but are in reality hard to reuse. For example, finding an implementation with an unnecessary ``craft sub-program'' may satisfy a given goal, but the requirement to craft the additional sub-program makes this implementation less desirable than an implementation of minimal length. As the library grows, the decompositions that are more reusable lead to more success executions, and thus have higher frequencies.}

\paragraph{Recency score}{This is a linear mapping from 0 to 1, from the oldest to the newest decomposition, respectively, among a set of candidate decompositions. Decompositions from the same session have a high likelihood of success because they were executed in similar contexts (e.g., using the same recipe, even if the inventory has changed slightly), and thus prioritizing recent decompositions makes the planner focus the search on decompositions with a high likelihood of success.}

\begin{lstlisting}[language=Python, caption={Inner propose}, label={lst:inner_propose}]
def inner_propose(context, np, library, parameters):
  # Look-up decompositions of the given signature 
  known_decompositions = get_decompositions(library, np.goal)

  distribution = [];
  for decomposition in known_decompositions:
    # Ignore decompositions that induce recursion
    if has_recursion(context, np, decomposition):
      continue

    # Compute the score of the decomposition
    recency_score = get_recency_score(decomposition)
    frequency_score = get_frequency_score(decomposition)
    score = parameters.recency_weight*recency_score +\
                  parameters.frequency_weight*frequency_score
    distribution.push([decomposition, score])

  # Sort by score
  distribution.sort()
  return distribution
\end{lstlisting}

\subsection{Outer propose}

The outer propose leverages the context and the natural language hint to propose programs in the library cross product. We experimented with two approaches: one based on enumeration and the other using Large Language models. We now describe these approaches in detail.

\paragraph{naive} We assign each program in the library with a recency score, ranging from 0 (oldest) to 1 (newest). The programs are sorted, and the top 12 are chosen, together with the primitives. This forms a beam which will be used to enumerate the cross-product space in lexicographic order, attempting to execute each partial NP plan with the usual NP semantics. The cross-product enumeration thus corresponds to the output of the \textit{propose} function.

\paragraph{distance} We use the SentenceTransformers library~\cite{reimers-2019-sentence-bert} to assign semantic similarity between the hints of natural programs in the library and the input program. This score is then added to the recency score, and the set is sorted and enumerated similarly to the \textbf{naive} proposer.

\paragraph{distance + gpt-3.5-turbo} This runs the same logic as \textbf{distance}, but while the interpreter is running we query in parallel gpt-3.5-turbo from the OpenAI API, and sample top-level decompositions. We then try executing each sampled top-level decomposition (note that this does not call the outer propose recursively) and return as soon as either process finds a solution. See below for prompt details.

\paragraph{distance + text-davinci-003} Similar to \textbf{distance + gpt-3.5-turbo}, except with text-davinci-003 as the LLM.

\subsection{Propose experiment}

\begin{figure*}[t]
\begin{center}
\includegraphics[width=0.48\textwidth]{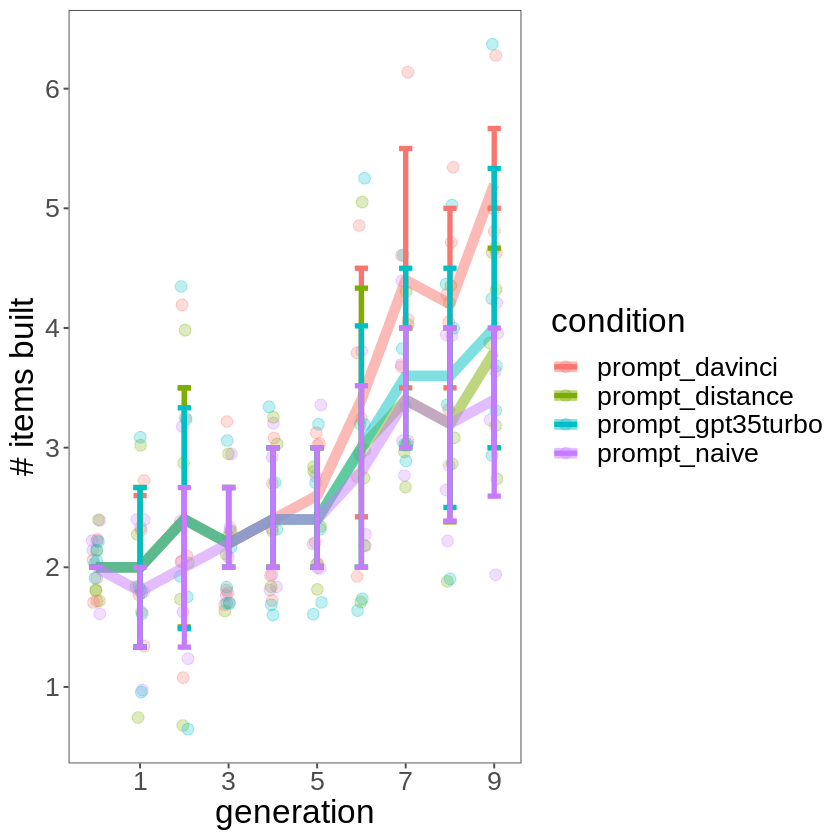}
\hfill
\includegraphics[width=0.48\textwidth]{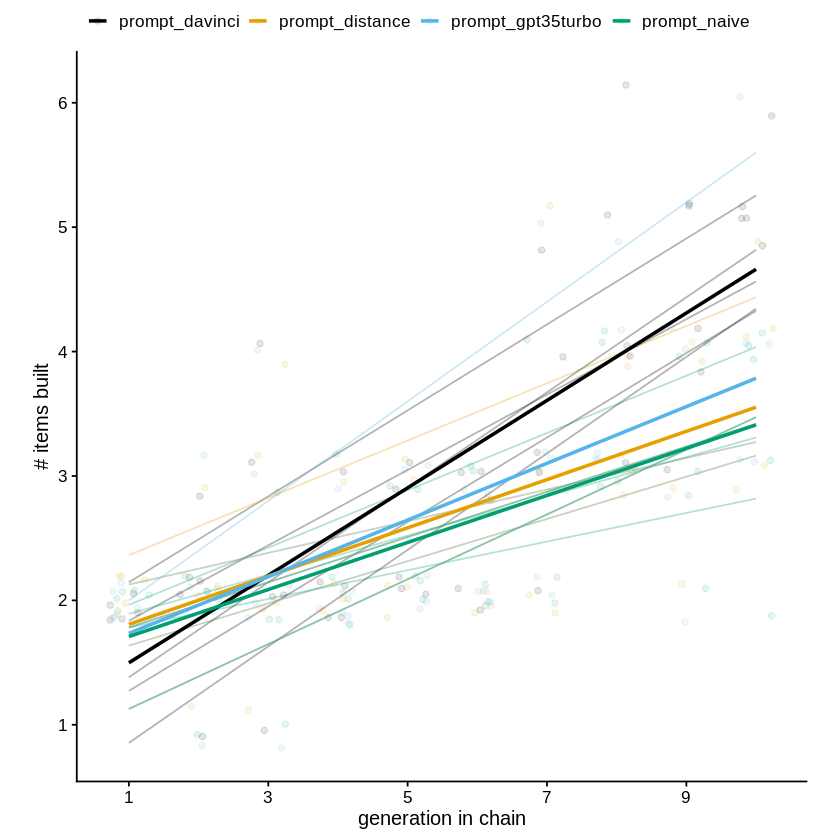}
\caption{Simulation results comparing performance for different outer propose implementations for NP. Overall, all outer propose implementations have similar performance, except for \textbf{distance+text-davinci-033} (here labeled ``prompt\_davinci''), which performs best in terms of items built.}
\label{fig:propose_plots}
\end{center}
\end{figure*}

\begin{figure*}[t]
\begin{center}
\includegraphics[width=0.48\textwidth]{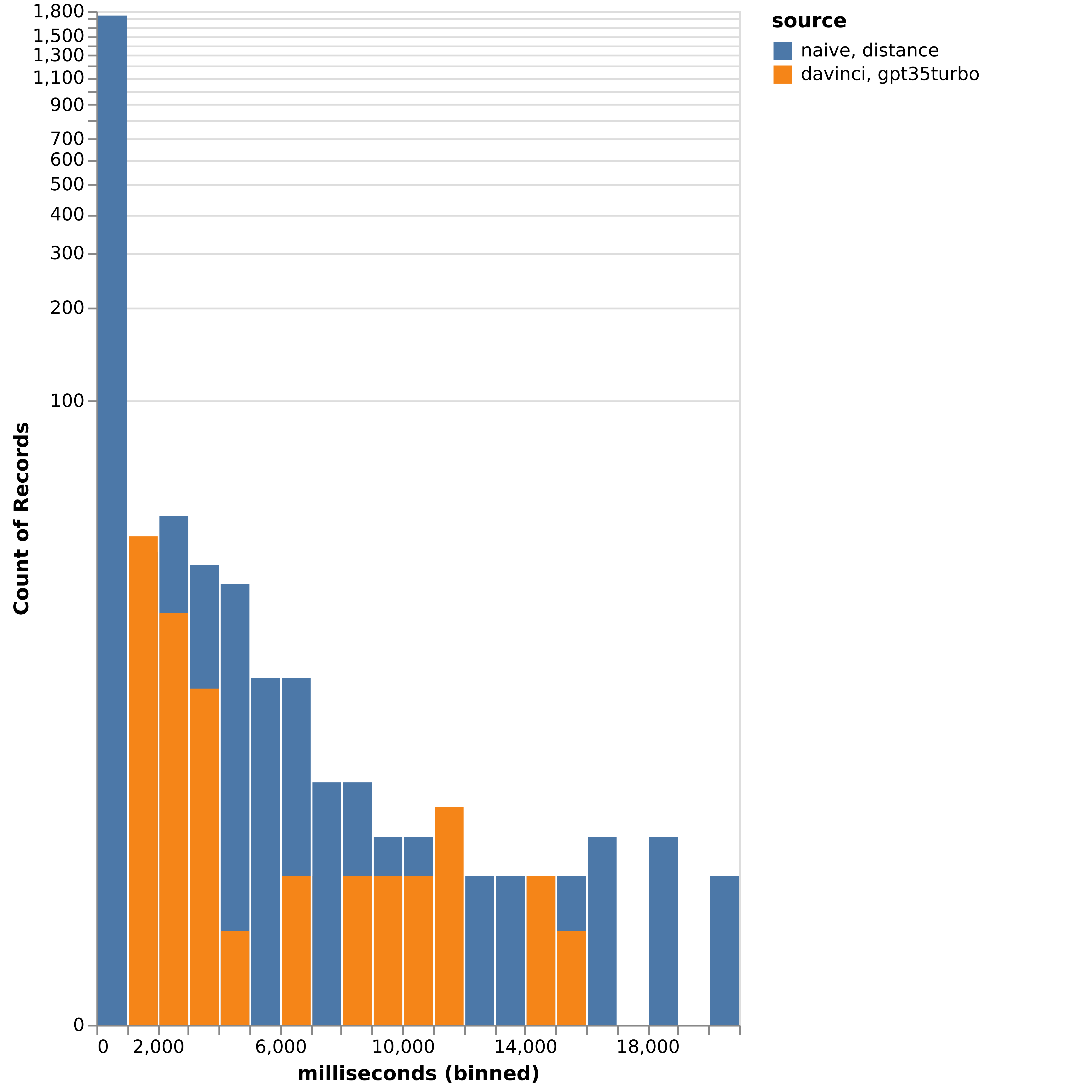}
\caption{Histogram counting the number of unique successful executions as a function of planner time according to the outer propose strategy (symlog scale on y-axis). Note that non-LLM strategies (\textbf{naive}, \textbf{distance}) are responsible for over an order of magnitude more successful executions. This figure omits executions which result in a program which is already in the library, as these are very efficiently handled by the inner propose in tens of milliseconds and thus require no outer propose considerations.}
\label{fig:solver_times}
\end{center}
\end{figure*}

To compare the different outer propose implementations, we set a timeout of 20 seconds and run the simulation study on 5 batches on all outer propose implementations for 10 generations with 50\% randomness. The better performing outer propose is \textbf{distance+text-davinci-003}, followed by \textbf{distance+gpt-3.5-turbo}, \textbf{distance} and \textbf{naive}, in order (see Figure~\ref{fig:propose_plots}) and with the latter three with similar performance. We also measured the time it takes the LLM and non-LLM propose functions to get to the correct answer, with the non-LLM propose functions being faster and being responsible for an order of magnitude more successful executions (see Figure~\ref{fig:solver_times}).

In pilot studies, we observed that users very rarely wait enough time for the LLM to be sampled. Additionally, we observed that \textbf{distance} seems to work best with real user inputs.
Because of these considerations, we decided to use the \textbf{distance} outer propose implementation, as that offers a good tradeoff between speed and performance.

However, we believe that LLMs can benefit \textit{propose} implementations more substantially in tasks with the following properties: (1) users willing to wait longer for the language model to be sampled, (2) tasks which require more complex structures, such as those involving control-flow statements, (3) local language models with reduced latencies.

\section{Prompt examples}
\label{sec:prompt}

We include an example prompt for each of LLMs used for reference, from an NP session. 
In both cases the prompt is built by providing the list of primitives and post conditions that the model can use, and by including examples from the library. The examples are chosen by selecting items in the library with semantic similarity. We use the SentenceTransformers library~\citep{reimers-2019-sentence-bert} to compute semantic similarity.

In both NP and DS, the returned completions are parsed and executed by the DS and NP systems as the input programs. In the case if \textbf{distance+llm} propose implementations, the \textbf{distance} propose and the LLM sampling are executed in parallel.

\subsection{text-davinci-003}
\begin{lstlisting}[mathescape=true,
    caption={Example prompt for text-davinci-003}, label={alg:promptdavinci},
    basicstyle=\small, %or \small or \footnotesize etc.,
    breaklines=true,
]
START
{"name":"please craft 'brick' with 'clay' and 'clay'","post_condition":"[\"brick\"]"}
    {"post_condition":"[\"sand\"]"}
    {"post_condition":"[\"clay\"]"}
    place clay
    place clay
    collect
END
START
{"name":"please craft 'clay' with 'sand' and 'pickaxe'","post_condition":"[\"clay\"]"}
    {"post_condition":"[\"pickaxe\"]"}
    place pickaxe
    place sand
    collect
END
START
{"name":"please craft 'clay' with 'sand' and 'pickaxe'","post_condition":"[\"clay\"]"}
    place wood_stick
    place stone
    collect
    {"post_condition":"[\"sand\"]"}
    {"post_condition":"[\"sand\"]"}
    place pickaxe
    place sand
    collect
END
START
{"name":"please craft 'clay' with 'sand' and 'pickaxe'","post_condition":"[\"clay\"]"}
    {"post_condition":"[\"pickaxe\"]"}
    {"post_condition":"[\"sand\"]"}
    place pickaxe
    place sand
    collect
END
START
{"name":"please craft 'clay' with 'sand' and 'pickaxe'","post_condition":"[\"clay\"]"}
    {"post_condition":"[\"sand\"]"}
    place pickaxe
    place sand
    collect
END
START
{"name":"please craft 'hut' with 'string' and 'grass'","post_condition":"[\"hut\"]"}
    place grass
    place string
    collect
END
START
{"name":"please craft 'hut' with 'string' and 'grass'","post_condition":"[\"hut\"]"}
    {"post_condition":"[\"string\"]"}
    place grass
    place string
    collect
END
START
{"name":"please craft 'hut' with 'string' and 'grass'","post_condition":"[\"hut\"]"}
    {"post_condition":"[\"string\"]"}
    place grass
    place string
    collect
END
START
{"name":"please craft 'hut' with 'string' and 'grass'","post_condition":"[\"hut\"]"}
    {"post_condition":"[\"string\"]"}
    place grass
    place string
    collect
END
START
{"name":"please craft 'hut' with 'string' and 'grass'","post_condition":"[\"hut\"]"}
    {"post_condition":"[\"string\"]"}
    place grass
    place string
    collect
END
START
{"name":"make clock with gears and cables","post_condition":"[\"clock\"]"}
\end{lstlisting}

\subsection{gpt-3.5-turbo}
\begin{lstlisting}[mathescape=true,
    caption={Example prompt for gpt-3.5-turbo}, label={alg:prompt},
    basicstyle=\small, %or \small or \footnotesize etc.,
    breaklines=true,
]
The task is to convert a sentence of how to craft an item into a small program that crafts the item.

Here are all the primitive functions that you can use:
["place wood","place stone","place wool","place grass","place wood_plank","place bed","place wood_stick","place paper","place string","place pickaxe","place cloth","place pinwheel","place hut","place stone_mill","place flour","place book","place fire","place oven","place bread","place sand","place clay","place brick","place glass","place house","place iron","place gear","place wire","place light_bulb","place clock","collect","clear"]

When you are placing something, you can only use the functions in this list.

Here are all the post conditions that you can use:
["[\"bed\"]","[\"wood_plank\"]","[\"wood_stick\"]","[\"string\"]","[\"cloth\"]","[\"paper\"]","[\"pinwheel\"]","[\"hut\"]","[\"stone_mill\"]","[\"flour\"]","[\"fire\"]","[\"oven\"]","[\"bread\"]","[\"book\"]","[\"string\"]","[\"pickaxe\"]","[\"flour\"]","[\"bread\"]","[\"hut\"]","[\"sand\"]","[\"clay\"]","[\"brick\"]","[\"house\"]","[\"fire\"]","[\"glass\"]","[\"light_bulb\"]","[\"bed\"]","[\"pinwheel\"]","[\"oven\"]","[\"sand\"]","[\"clay\"]","[\"brick\"]","[\"glass\"]","[\"cloth\"]","[\"paper\"]","[\"stone_mill\"]","[\"iron\"]","[\"wire\"]","[\"light_bulb\"]","[\"iron\"]","[\"gear\"]","[\"clock\"]","[\"wood_plank\"]","[\"wire\"]","[\"clock\"]","[\"gear\"]","[\"wood_stick\"]","[\"pickaxe\"]","[\"book\"]","[\"house\"]","[\"clock\"]"]

Here are some examples:
Input:
{"name":"please craft 'clock' with 'gear' and 'wire'","post_condition":"[\"clock\"]"}
Output:
[
    {"post_condition":"[\"wire\"]"}
    {"post_condition":"[\"gear\"]"}
    place gear
    place wire
    collect
]

# more examples...

New input:
{"name":"please craft clock with 'gear' and 'book'","post_condition":"[\"clock\"]"}
What is the output?
Give five output examples for this new input with post conditions and five without.
Your examples MUST be for this specific new input.
\end{lstlisting}
\section{NP execution}
\label{sec:np_execution}

We first provide implementation details of lgorithm~\ref{alg:NP} and then further describe how the NP system learns from previous experiences.

In our implementation, we explicitly enumerate the library cross-product at the ``top-level'' of the recursion, instead of relying on rejection sampling, which is very inefficient. That is, we implement an outer loop where sequences of NPs are enumerated and an inner loop with the traditional semantics. This section instead describes the ``inner'' search, which employs the inner propose.

\paragraph{Implementation}{
Our implementation employs an explicit recursion stack with all the search trees under consideration. To avoid redundant execution of partial plans we cache the ``partial execution'' of each plan. Initially, there is only one plan in the stack corresponding to the input Natural Program. This plan is popped and partially executed, at which point we call the inner propose to obtain decompositions. The plan is then expanded with each decomposition, resulting in new partial plans which are then added to the stack. This process is then repeated until the stack is empty or a solution is found (or the planning timeout is reached).
}

\paragraph{Learning from experiences}{
Successful execution a Natural Program results in a plan tree containing all the decompositions used and the Natural Programs to which they belong, eventually resulting in decompositions with only primitives. To learn from a library of such trees, an NP system recursively extracts all the decompositions in the trees and stores them in a data structure which allows efficient reuse in later searches (see Inner propose section). This data structure is effectively a mapping from goals to top-level decompositions associated with a given goal. Thus, this allows the planner to leverage all previously known decompositions when executing a new Natural Program.
}

\begin{lstlisting}[language=Python, caption={NP interpreter pseudocode. GPC semantics are bound to the \texttt{partial\_eval} function.}, label={lst:np_code_queue}]
def run_np(np, library, partial_eval):
  # Build depth-first search stack
  stack = Stack([np])

  # Run best-first search by popping from the queue
  while stack.size() > 0:
    top_candidate = stack.pop()

    # Resume execution of top candidate
    result = partial_eval(top_candidate)

    if "last_state" in result:
      # Case 1:
      # All constraints were successful.
      return result
    else if not "error" in result:
      # Case 2:
      # The interpreter found an un-expanded NP ("signature"),
      signature = result.signature

      # Use proposer to sample replacements for the signature
      replacements = proposer(signature, library)

      # Get new candidates by performing the replacements
      # on the annotated program.
      for replacement_np in replacements:
        new_np = get_replaced(
          result,  # search in partial program
          signature,  # search for unexpanded NP
          replacement_np,  # replace with proposed expanded NP
        )
        stack.push(new_np)
    else if "error" in result:
      # Case 3:
      # The program crashed (maybe because a goal failed)
      # There is nothing to do. Continue with next candidate.
      pass
  raise error
\end{lstlisting}

\section{Workflow}

In this section we provide details on the human experiment workflow design.

\begin{figure*}
\begin{center}
\includegraphics[width=0.48\textwidth]{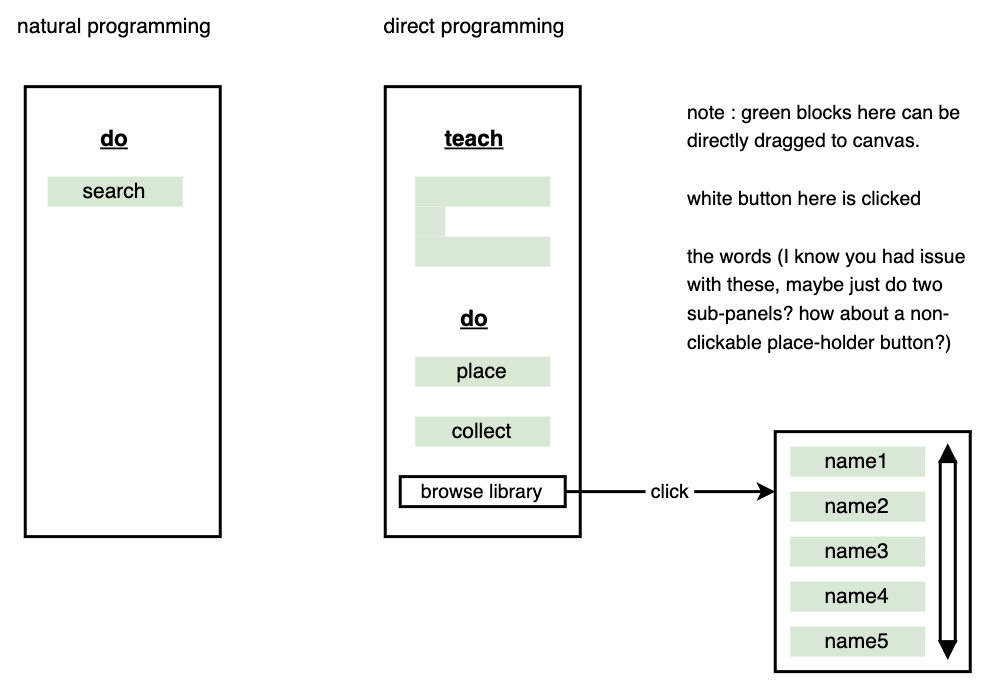}
\caption{Left panel in Natural Programming (and Direct Synthesis) and Direct Programming. In the final UI, the ``brows library'' button opens a modal with search bar which filters the library to the programs with a hint containing the query as a substring.}
\label{fig:workflow_panel_np_ds_dp}
\end{center}
\end{figure*}

\begin{figure*}
\begin{center}
\includegraphics[width=0.48\textwidth]{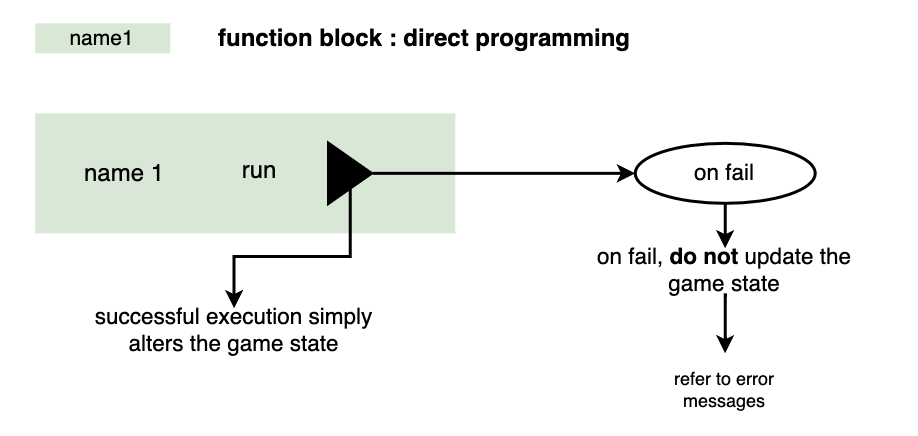}
\caption{UI semantics of executing a program in Direct Programming.}
\label{fig:workflow_function_dp}
\end{center}
\end{figure*}

\begin{figure*}
\begin{center}
\includegraphics[width=0.48\textwidth]{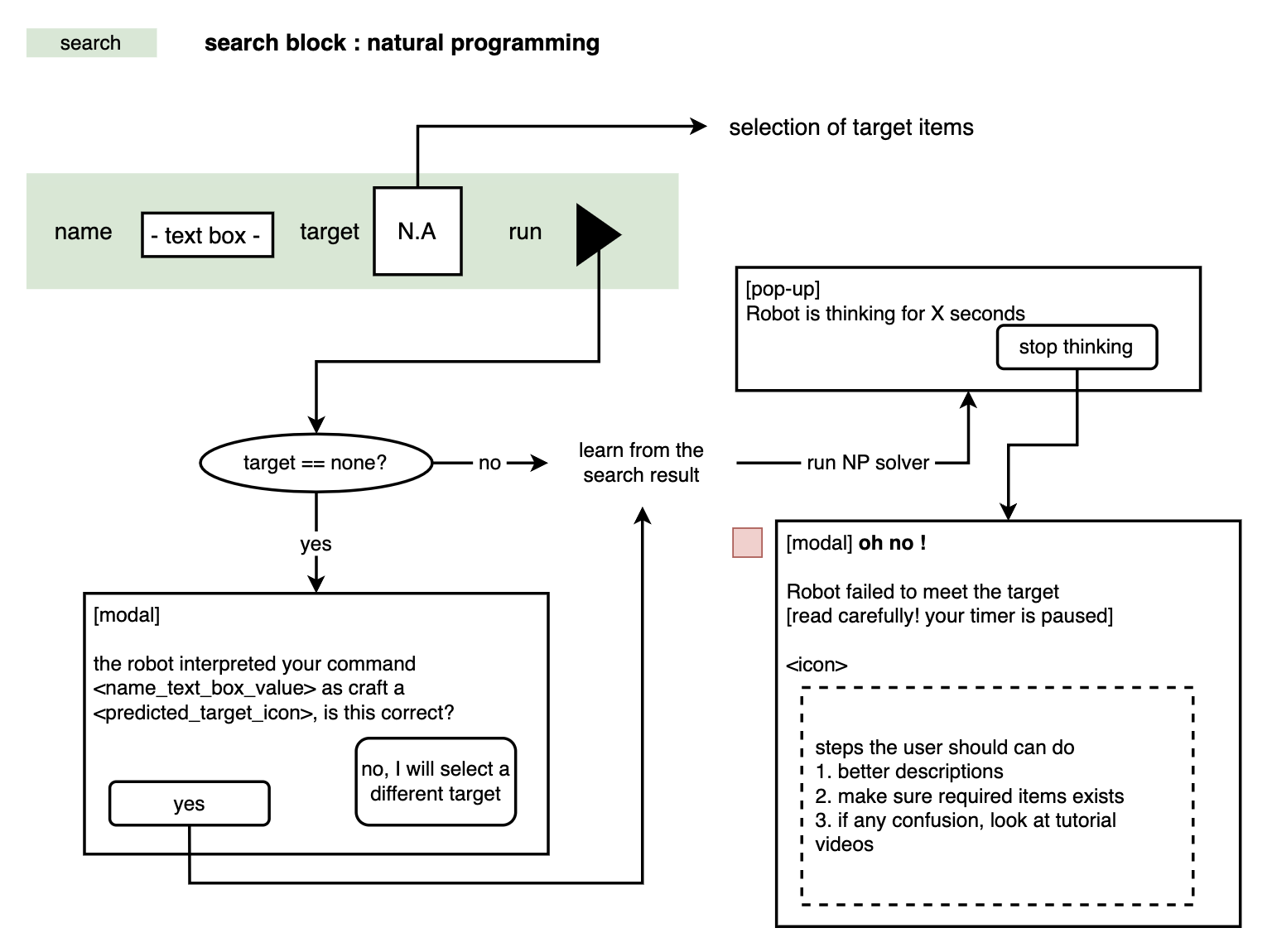}
\caption{UI semantics of executing a program in Natural Programming and Direct Synthesis. If a program without a goal is executed, semantic search is used to identify the item in the library with most similar name, and the goal in that program is suggested to the user (we use the SentenceTransformers library~\citep{reimers-2019-sentence-bert} to compute semantic similarity). If the user confirms the goal, then it is added to the program and the planner is called, otherwise, the user is returned to the main UI to manually set a goal from a dropdown menu.}
\label{fig:workflow_search_block_np_ds}
\end{center}
\end{figure*}

\begin{figure*}
\begin{center}
\includegraphics[width=0.48\textwidth]{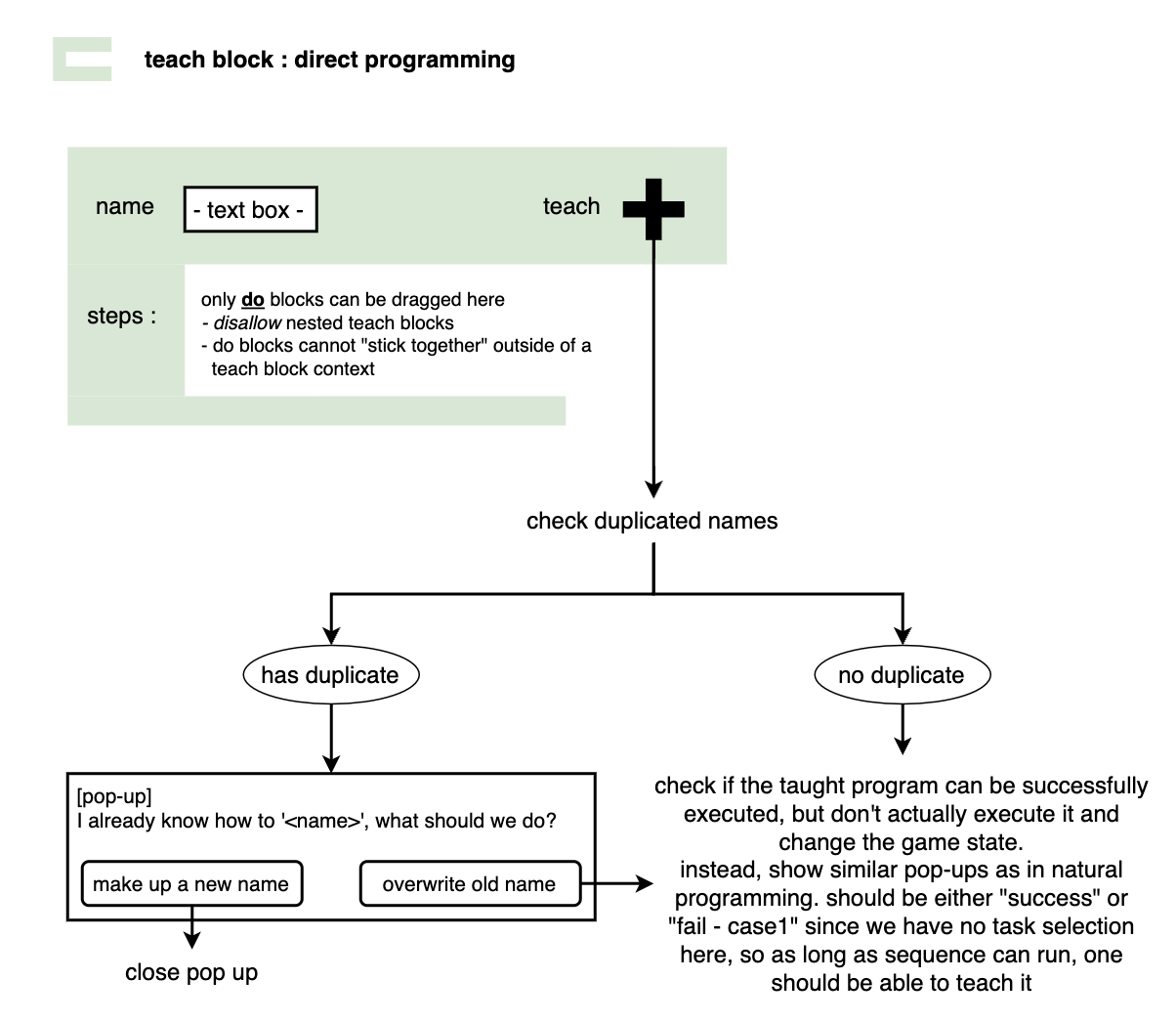}
\caption{``Teach'' semantics in Direct Programming. The teach button saves programs into the library if the program has a unique identifier. This constraint follows from the fact that Direct Programming semantics cannot disambiguate between competing implementations.}
\label{fig:workflow_teach_block_dp}
\end{center}
\end{figure*}

\paragraph{Workflow diagrams}{
These diagrams describe the flow of interactions that happen as the user engages with the \CL{} user interface. Figure~\ref{fig:workflow_panel_np_ds_dp} shows the left panel (``toolbox'' in Blockly) as well as a short description of their function. Figure~\ref{fig:workflow_function_dp} shows the UI semantics of executing a program in the Direct Programming mode. Figure~\ref{fig:workflow_search_block_np_ds} shows the UI semantics of executing a program in Natural Programming and Direct Synthesis. Figure~\ref{fig:workflow_teach_block_dp} shows the UI semantics of adding a program to the library in Direct Programming.
}

\begin{figure*}
\begin{center}
\includegraphics[width=0.48\textwidth]{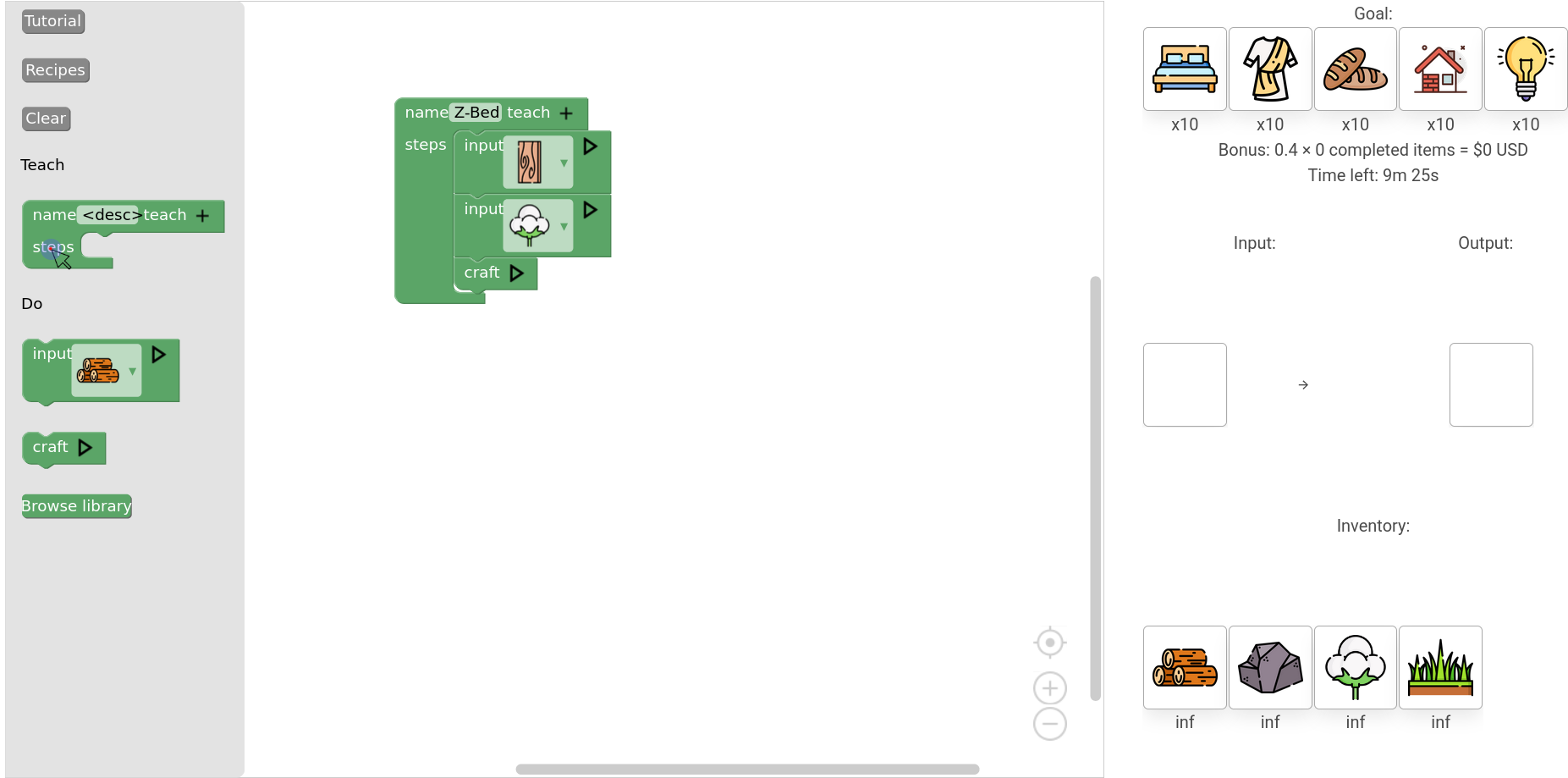}
\caption{User interface for Direct Programming. Users are provided an interactive workspace to write programs and visualize the library. Programs are written by dragging a ``C-block'' unto the canvas. This block has a field to specify a program name, and slots to which instruction blocks can be added. Users can add programs from the library to the workspace, as well as primitive functions. The interface also includes the current program state, the session goals and the session timer.}
\label{fig:ui_dp}
\end{center}
\end{figure*}

\begin{figure*}
\begin{center}
\includegraphics[width=0.48\textwidth]{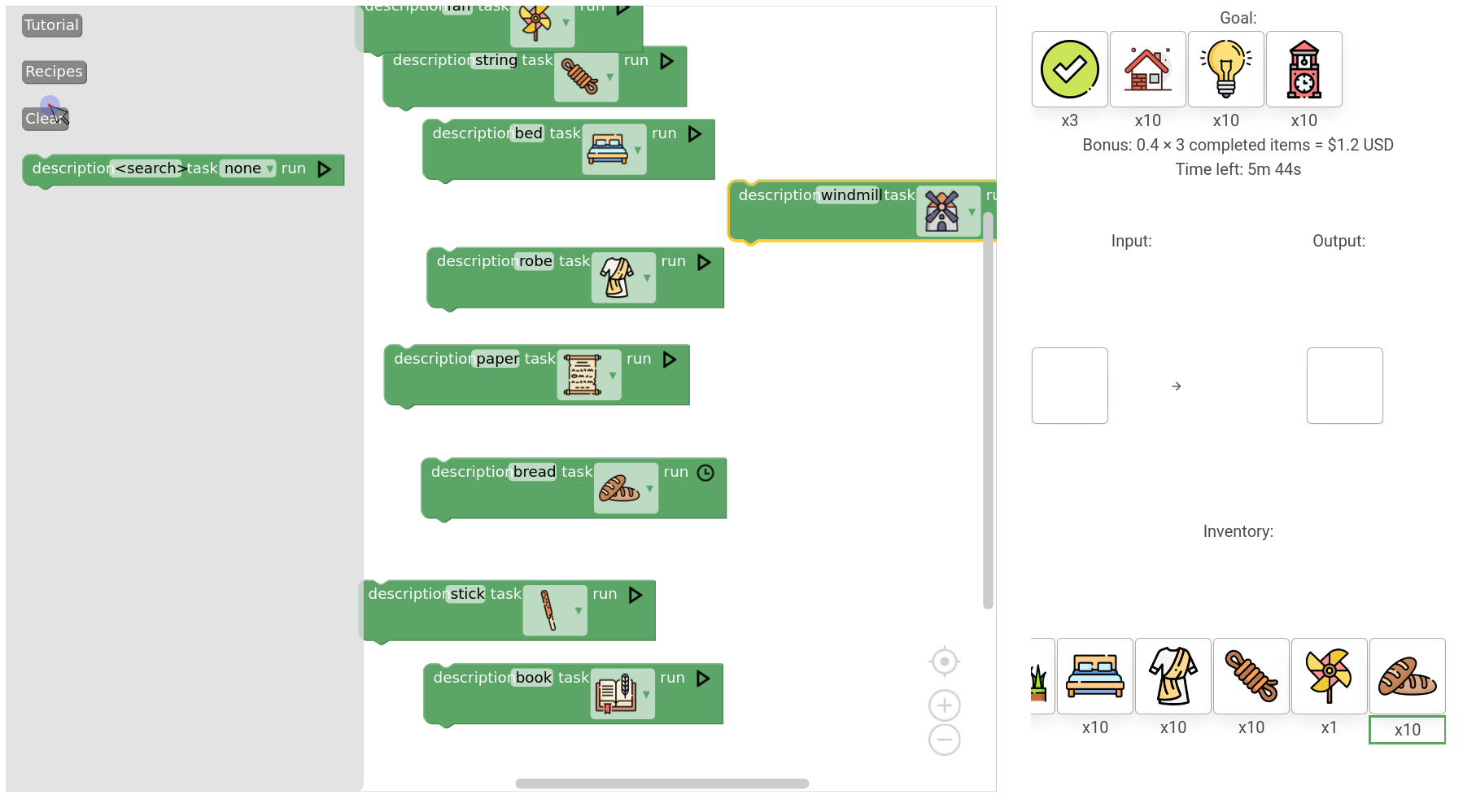}
\caption{User interface for Natural Programming and Direct Synthesis. Users are provided an interactive interface to describe search problems with natural language, and to specify goals with pictures if necessary. Search problems are described with blocks, which the user drags onto a workspace. Users can move, remove and reuse these blocks as they see fit. The user interface also includes the current program state, the session goals and the session timer.}
\label{fig:ui_np_ds}
\end{center}
\end{figure*}

\paragraph{Blockly UI}{
The UI was implemented with Blockly~\citep{8120404}, as it provides an intuitive and interactive framework for programmatic tasks. The UI revolves around a workspace, to which blocks can be dragged. In the case of Direct Programming, blocks support defining functions and browsing the library (see Figure~\ref{fig:ui_dp}). In the case of Natural Programming and Direct Synthesis, blocks are used to specify search problems (see Figure~\ref{fig:ui_np_ds}).
}

\begin{figure*}
\begin{center}
\includegraphics[width=0.48\textwidth]{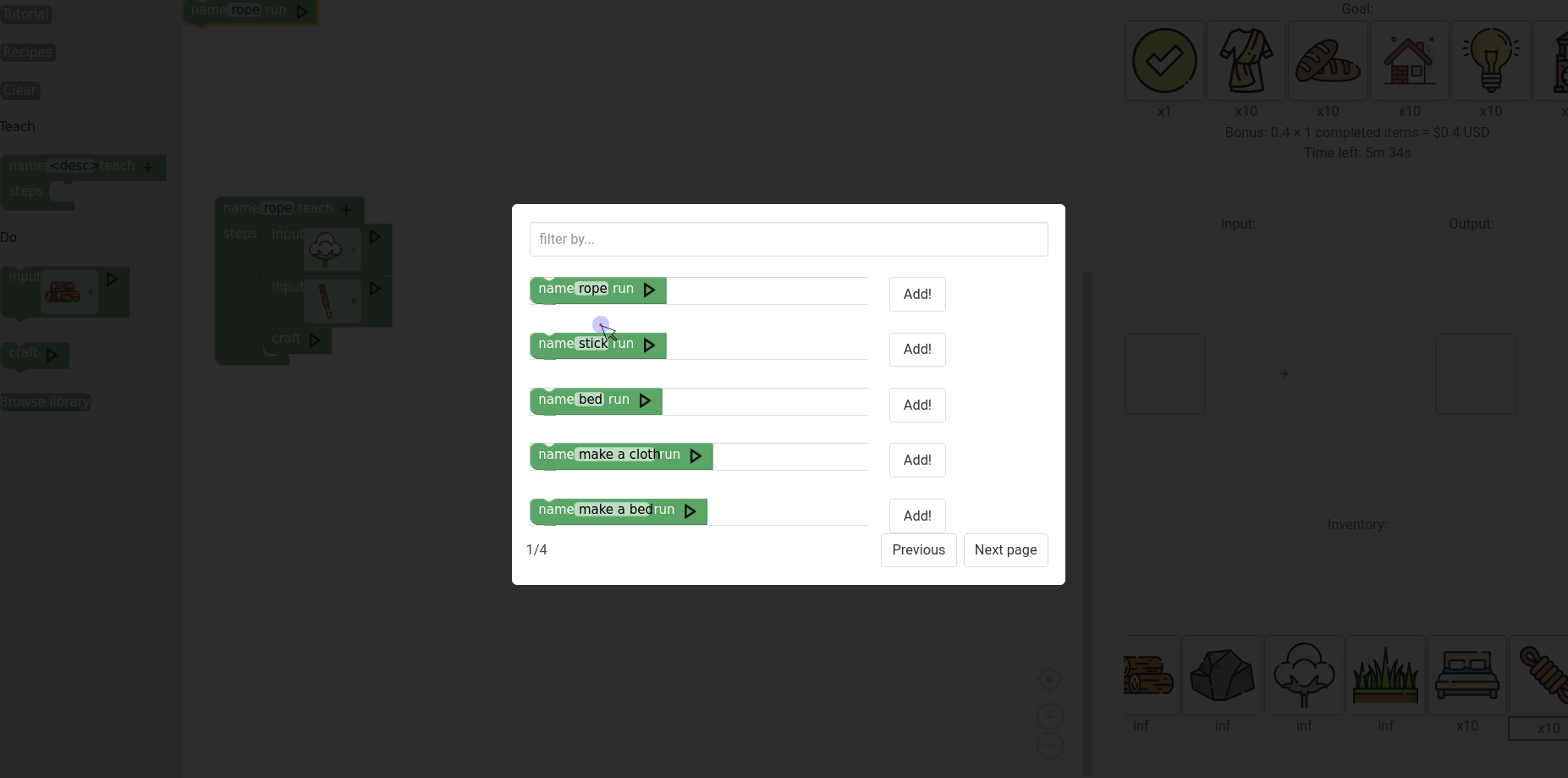}
\caption{User interface for visualizing the library. In the Direct Programming mode, users can browse, filter and add programs in the library to the workspace.}
\label{fig:ui_library}
\end{center}
\end{figure*}

\paragraph{Library UI}{
All three systems --Direct Programming, Natural Programming and Direct Synthesis-- maintain a library. However, explicit browsing of the library is only necessary in the case of Direct Programming, as the other systems can leverage the library automatically without user intervention. The user interface for browsing the library allows users to see and filter programs in the library. Every program can be added to the workspace by clicking a button next to it. Filtering is done through a text input, which filters the list of programs to those that contain the query as a substring in their name. See Figure~\ref{fig:ui_library}.
}

\end{document}